\documentclass{SPAICE}
\def\authorEmail{tat-jun.chin@adelaide.edu.au}
\def\AuthorShort{H. M. Nguyen, M. Märtens and TJ Chin}

\author[1]{Hong Minh Nguyen}
\author[1]{Marcus Märtens}
\author[1]{Tat-Jun Chin\thanks{Corresponding author. E-Mail: \authorEmail}}
\affil[1]{AI for Space Group, Adelaide University, Australia}

\title{Learning Cross-view Correspondences for Geo-localization on Planetary Surfaces}

\usepackage{graphicx}
\usepackage{subcaption}
\usepackage{float}
\usepackage{xcolor}
\usepackage{hyperref}
\begin{document}
\raggedbottom
\maketitle

\begin{abstract}
Maintaining global position awareness is a fundamental challenge for planetary surface exploration, since satellite-based positioning systems are unavailable and onboard odometry drifts over time.
Although orbital mapping products, such as overhead imagery and terrain-derived maps, provide global context, aligning them with surface observations is challenging due to large viewpoint differences, low texture, repetitive terrain, and drastic changes in appearance caused by varying illumination and topography. 
We introduce a new cross-view geo-localization benchmark built from physically rendered surface panoramas and overhead tiles derived from a high-resolution lunar terrain model.  
Our dataset contains 10\,438 ground views rendered as 360$^\circ$ surface panoramas with matching overhead images precisely centered at the same location.
Additionally, a set of overlapping tiles is provided to study off-center localization with multiple plausible candidates per panorama.
We study the performance of a state-of-the-art transformer-based geo-localization method on our data, by training it from scratch and reporting retrieval accuracy.
Our results demonstrate that learning-based cross-view localization methods can be successfully applied to the domain of planetary surfaces, providing a vision-based alternative to global navigation satellite systems.
\end{abstract}

\section{Introduction}

Accurate localization is a core requirement for autonomous exploration of extraterrestrial planetary bodies, where Global Navigation Satellite Systems (GNSS) are not available~\cite{Chen2024}. Surface vehicles must therefore maintain global position awareness despite drift in locally estimated motion and pose over time. This motivates the use of orbital mapping products, such as overhead imagery and terrain-derived representations, as a consistent map-like reference over large areas.~\cite{https://doi.org/10.1029/2010GL043751}. However, exploiting these products for localization requires establishing reliable correspondences between surface observations and overhead views, despite large viewpoint changes and appearance variation caused by illumination and terrain geometry~\cite{GLASER201478}. 

Cross-view localization addresses this correspondence problem by relating ground-level observations to overhead, map-like representations. Several formulations are possible, including retrieval-based matching, direct pose regression, and probabilistic alignment against a map~\cite{shi2020beyond}. For this work, we study cross-view localization as an image retrieval task consistent with widely used terrestrial benchmarks such as CVUSA~\cite{workman2015wide}, CVACT~\cite{Liu_2019_CVPR}, and VIGOR~\cite{zhu2021vigor}. While most prior work on cross-view localization focuses on urban scenes, our research question is to what extent machine learning methods can be applied to non-terrestrial planetary surfaces?

In contrast to the urban environments of Earth, the natural terrains of other planets, moons and asteroids offer only sparse texture, minimal semantic cues, and weak structural priors, as their geomorphology is dominated by slopes, isolated rocks, and impact craters.
Many of those features vary heavily in their appearance as the sun illumination can produce deep shadows and bright reflections on planetary terrain~\cite{craterlight}.
When ground and overhead views are not temporally aligned, such illumination effects can lead to substantial inconsistency between the two viewpoints.
Computer-generated images have become a practical way to study cross-view correspondences under controlled conditions. Previous studies such as NASA JPL's LunarNav~\cite{daftry2023lunarnav} or Zhao et al.~\cite{10681493} rely on synthetic terrain renderings using Blender\footnote{\url{https://www.blender.org}} while exploring a rover localization problem, often harnessing crater detection techniques~\cite{matthies2022lunar}.

Our contribution differs from those previous works in two key aspects: Firstly, we render our views using PANGU~\cite{pangupaper}, an established planetary surface simulator that has been cross-validated with space missions for high realism. Secondly, we do not assume a rover setting, nor do we explicitly enforce reliance on the presence of craters or other geometries that might be specific to lunar terrain only. 
Instead, our dataset follows the structural design and organization of established terrestrial benchmarks like CVUSA and VIGOR, allowing for a direct transfer of the advanced methods that have been developed for them by the computer vision community, while replacing their urban scenes with planetary surface imagery such as lunar terrain.
To demonstrate transferability, we train TransGeo~\cite{zhu2022transgeo}, a state-of-the-art cross-view localization model, on our dataset from scratch. Analyzing the performance of TransGeo allows us to extract practical insights related to the training of cross-view localization models and their robustness against inconsistent illumination. On a more general level, our dataset enables the study of visual cues that drive cross-view correspondence generation within repetitive planetary terrains, providing a foundation for the development of future methods aiming to close the gap between simulated data and real planetary surface imagery.

\section{Dataset}
\subsection{Location Selection and Tiling}
\label{sec:sampling}
We base our cross-view data on LROC/NAC~\cite{robinson2010lroc}, a high-resolution digital terrain model (DTM) of the Moon. Several points of interest (POIs) are sampled on a regular grid over this terrain, with adjacent grid points spaced 110 m apart before random jitter is applied to their locations. For each POI, we render 16 evenly spaced perspective views with fixed camera parameters and stitch them together into a 360$^\circ$ surface panorama using cylindrical projection with the POI at the panorama center and a consistent absolute orientation.
We limit our sampling to the interior region of the DTM to avoid edge artifacts, covering approximately 130.5~km$^2$ and resulting in a set of 10,438 ground panoramas in total.

We investigate two variants for the association of overhead imagery to panoramas, inspired by established urban cross-view benchmarks:
\begin{enumerate}
    \item \textbf{One-to-One (CVUSA-style)}: Each panorama is paired with a single overhead image rendered with its center aligned to the POI.
    \item \textbf{Tile-based (VIGOR-style)}: Overhead images are rendered independently from the panorama POIs on positions of a regular grid covering the same geographical region. Afterwards, each panorama is associated with all the tiles that contain its POI.
\end{enumerate}

\begin{figure}[!ht]
  \centering

    \begin{subfigure}{\linewidth}
    \centering
    \includegraphics[width=\linewidth]{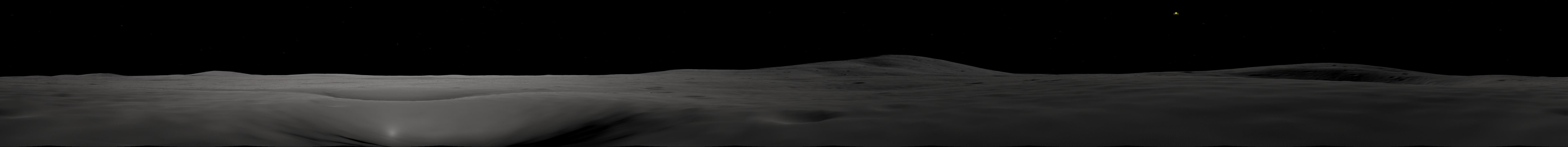}
    \caption{Ground panorama}
    \label{fig:q_ground_1}
  \end{subfigure}


  \begin{subfigure}{0.45\linewidth}
    \centering
    \includegraphics[width=\linewidth]{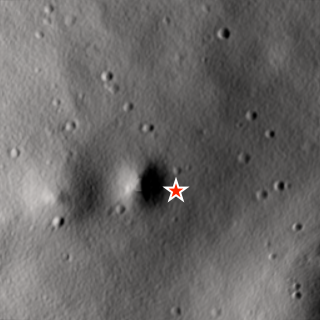}
    \caption{Positive tile.}
    \label{fig:q_pos}
  \end{subfigure}\hfill
  \begin{subfigure}{0.45\linewidth}
    \centering
    \includegraphics[width=\linewidth]{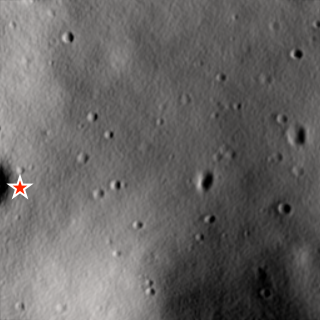}
    \caption{Semi-positive tile 1.}
    \label{fig:q_s1}
  \end{subfigure}


  \begin{subfigure}{0.45\linewidth}
    \centering
    \includegraphics[width=\linewidth]{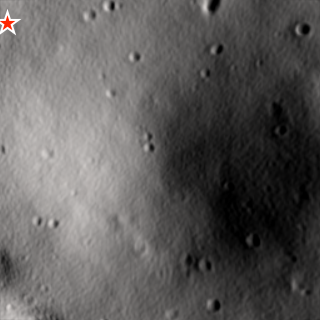}
    \caption{Semi-positive tile 2.}
    \label{fig:q_s2}
  \end{subfigure}\hfill
  \begin{subfigure}{0.45\linewidth}
    \centering
    \includegraphics[width=\linewidth]{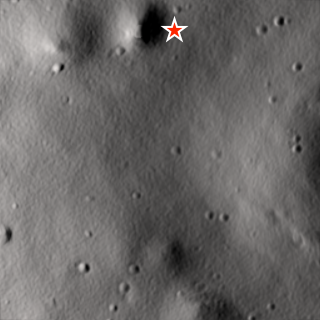}
    \caption{Semi-positive tile 3.}
    \label{fig:q_s3}
  \end{subfigure}

  \caption{Example data entry for the tile-based dataset variant with ground panorama (a), one positive (b) and three semi-positive tiles (c,d,e). We indicate the POI by a red star.}
  \label{fig:tile_based_example}
\end{figure}

The tile-based variant allows to study the more challenging off-center localization, where the the distance between the center-pixel and the panoramic POI is greater than zero. A \emph{positive} tile is defined as the overhead image whose center is closest to the POI. A \emph{semi-positive} tile does still include the POI, but close to the boundary region, representing a plausible but weaker match than the positive tile within an image retrieval setup. Fig.~\ref{fig:tile_based_example} shows an example of a panorama together with its associated positive and semi-positive overhead tiles.

\begin{figure}[!htb]
    \centering
    \includegraphics[width=0.62\linewidth]{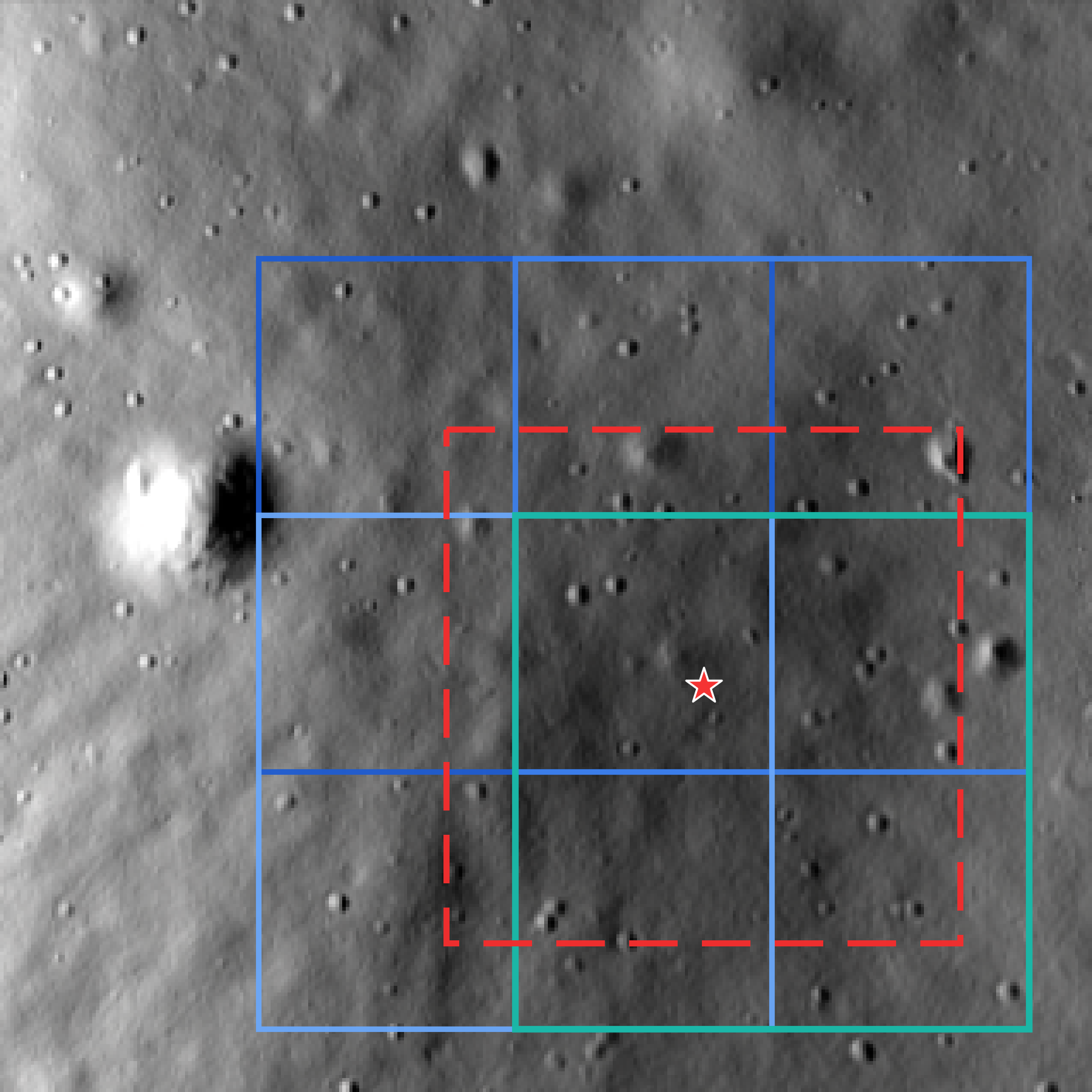}
    \caption{Tile-based sampling strategy. For a panorama at a POI \textcolor{red}{(red star)}, the aligned overhead tile is shown in dashed red. Of the four surrounding overlapping tiles, the closest is the \textcolor{teal}{positive tile}, and the other three are \textcolor{blue}{semi-positive tiles}.}
    \label{fig:tiling_strategy}
\end{figure}

In the tile-based variant, each panorama is always associated with one positive and three semi-positive tiles as shown in Fig.~\ref{fig:tiling_strategy}. However, due to the random sampling of the POIs, the number of panoramas associated with a single overhead tile can vary significantly. We will discuss the implications of this imbalance on data splitting and training in the later sections.

\subsection{Terrain Modeling and View Rendering}
The LROC/NAC DTM is imported into PANGU as the base terrain geometry. We enhance it using a high-frequency elevation overlay, additional high-resolution terrain layers, and synthetically generated craters. These geometric enhancements add small-scale detail while preserving the large-scale topography of the source DTM. Finally, we apply a texture derived from high-resolution LROC NAC imagery of the Apollo 11 region~\cite{HAPKE201675}.

\begin{figure}[!htbp]
    \centering
    \begin{subfigure}[t]{0.48\linewidth}
        \centering
        \includegraphics[width=\linewidth]{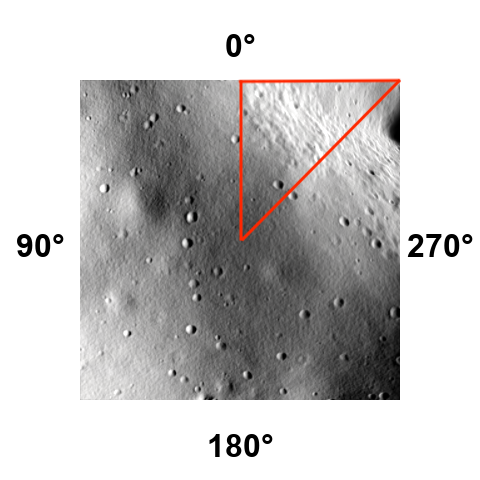}
        \caption{Orientation convention for the top-down tile, where angles are defined around the tile center. The red triangular sector indicates the viewing direction corresponding to the ground-level observation.}
    \end{subfigure}
    \hfill
    \begin{subfigure}[t]{0.48\linewidth}
        \centering
        \includegraphics[width=\linewidth]{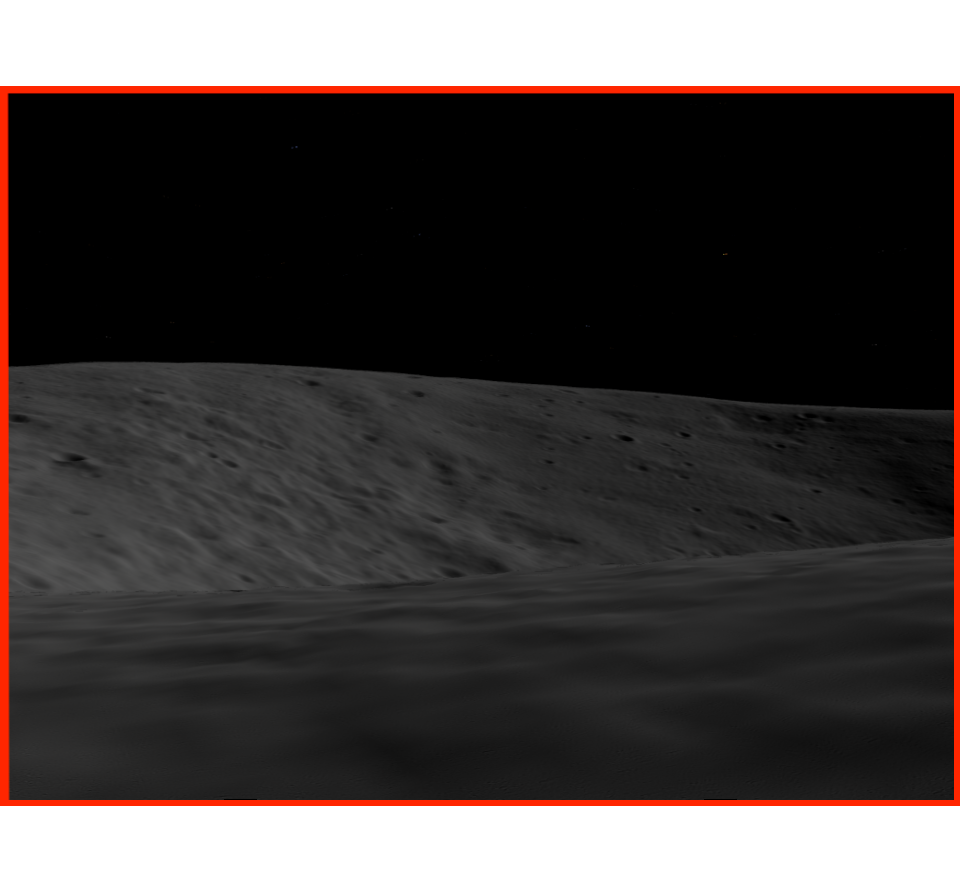}
        \caption{Zoom-in view of the ground panorama corresponding to the highlighted region in (a) and the selected region in (c)}
    \end{subfigure}

    \vspace{6pt}

    \begin{subfigure}{\linewidth}
        \centering
        \includegraphics[width=\linewidth]{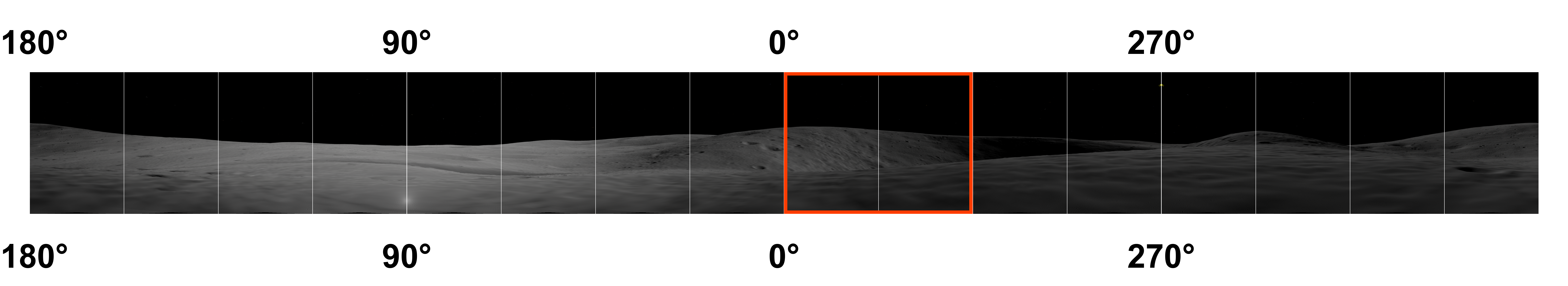}
        \caption{Example of 360$^\circ$ ground panorama. The red boxed region indicates the panorama segment whose viewpoint corresponds to the red sector shown in (a) and the ground view displayed in (b)}
    \end{subfigure}

    \caption{View geometry and orientation convention showing the relationship between the top-down tile orientation and the ground panorama viewpoint.}
    \label{fig:orientation1}
\end{figure}

All images generated in PANGU share the same rendering configuration across all variants of the data. The individual perspective views used to assemble the panoramas are rendered using the virtual camera positioned at a height of 7.5~m above the surface and with a field-of-view (FOV) of 60$^\circ$. Overhead imagery is rendered with a 90$^\circ$ FOV at a fixed camera height of 300~m. The sun azimuth and elevation are kept constant at 90$^\circ$ and 15$^\circ$ respectively, unless otherwise stated.
Fig.~\ref{fig:orientation1} illustrates the orientation of our views, which is kept consistent for all variants of our data. Each overhead image is rendered without any random in-plane rotation and aligned to a fixed world-frame reference (0$^\circ$ top, 90$^\circ$ left, 180$^\circ$ bottom, 270$^\circ$ right). The constituent 16 perspective views per panorama are taken at evenly spaced azimuth angles around the POI and stitched in azimuth order to form a 360$^\circ$ panorama. 
Under this convention, panorama columns correspond to increasing azimuth, and the panorama has a fixed starting direction aligned to the same 0$^\circ$ reference used in the overhead imagery.
As a result, there is no unknown rotational offset between the ground and overhead views. 
This allows models to focus on learning cross-view correspondences under viewpoint change without the additional challenge of resolving orientation ambiguity.

\section{Experiments} 

The purpose of our dataset is to enable the transfer of established cross-view localization techniques to the setting of planetary surfaces.
We select TransGeo~\cite{zhu2022transgeo} as the model to generate a proof of concept, as it is a modern transformer-based approach which is openly available\footnote{\url{https://github.com/Jeff-Zilence/TransGeo2022}} and whose performance and robustness have been demonstrated independently in several works~\cite{10.1007/978-3-031-72630-9_13, 10.1007/978-3-031-73021-4_3, 10769461, DBLP:conf/eccv/PillaiRS24} 

To prepare TransGeo for training, we split both dataset variants into train, validation and test sets, comprising 8\,354, 1\,031, and 1\,053 ground panoramas, respectively. To avoid information leakage, our splits are sampled from geographically separated areas of the DTM.

The architecture of our TransGeo implementation is the same as the original work, but with slight modifications to the training as follows: The model is trained by using AdamW with a learning rate of $3\times10^{-5}$, a weight decay of $0.03$, and cosine learning rate scheduling. Overhead images are resized to $320\times320$ for the tile-based variant, and $265\times265$ for the one-to-one  variant. The tile-based variant is trained for 60 epochs, while the one-to-one variant is trained for 40 epochs. 
Previous works (e.g.~\cite{deuser2023sample4geo}) have shown that the batching of positive and negative tiles has significant impact on the learning process and the final performance.
Consequently, we deploy and investigate the following data mining methods:

\begin{enumerate}
    \item \textbf{Data balancing only}: We sample up to six panoramas associated with each overhead tile to reduce overrepresentation of tiles with many panorama matches. No hard-negative mining is applied; eligible non-matching tiles are sampled randomly through mini-batch construction and used as in-batch negatives, without selection based on feature similarity or geographic proximity.
    
    \item \textbf{TransGeo mining}: In addition to data balancing, we use the global hard-negative mining scheme from the original TransGeo implementation. Candidate overhead tiles are ranked by their similarity to the panorama in the learned feature space, and highly similar non-matching tiles are used to construct the training batches. This selects feature-similar negatives regardless of their geographic distance.

    \item \textbf{Ring mining (proposed in this work)}: We sample negative tiles from a geographic ring around the panorama POI. These nearby tiles are likely to have similar terrain appearance, making them hard but valid negatives and encouraging the model to distinguish between neighbouring locations.
    
\end{enumerate}

To perform localization, TransGeo embeds the ground panorama and each overhead image candidate into a shared feature space. For a query embedding and a reference embedding, similarity is computed using cosine similarity and the reference images are ranked accordingly. The top-ranked overhead images are then returned and compared with the ground-truth positive and semi-positive tiles for evaluation. The Recall@$K$ metric (R@$K$) reports the fraction of queries for which the positive tile appears among the top $K$ retrieved candidates. The Hit@$K$ metric is less strict and treats semi-positive tiles as equally acceptable as positive tiles. Consequently, the Hit@$K$ metric only applies to the tile-based variant of our dataset.

\section{Results}
\label{sec:results}

\subsection{One-to-One (CVUSA-style) retrieval}
\label{sec:results_cvusa}

\begin{table}[t]
\centering
\caption{Retrieval accuracy of TransGeo for one-to-one (CVUSA-style) data under varying sun azimuth.}
\label{tab:results_cvusa_combined}

\small
\setlength{\tabcolsep}{4pt}
\renewcommand{\arraystretch}{1.15}

\begin{tabular}{lcccc}
\hline
Setting & R@1 & R@5 & R@10 & R@1\% \\
\hline
Matched illum. az=$90^\circ$, el=$15^\circ$
& 87.88 & 95.73 & 97.29 & 98.85 \\

Shifted illum. az=$270^\circ$, el=$15^\circ$
& 4.039 & 8.927 & 14.72 & 14.72 \\
\hline
\end{tabular}
\end{table}

Table~\ref{tab:results_cvusa_combined} shows the performance of TransGeo when applied to the One-to-One variant of our data. With a R@1 of 87.88\% this performance is only about~6\% worse than the reported R@1 of 94.08\% from the original TransGeo work~\cite{zhu2022transgeo} when trained on the original CVUSA dataset~\cite{workman2015wide}.
This highlights that TransGeo is capable of cross-localizing planetary terrain if conditions are ideal.

\begin{figure}[!htbp]
    \centering

    \begin{subfigure}{\linewidth}
        \centering
        \includegraphics[width=\linewidth]{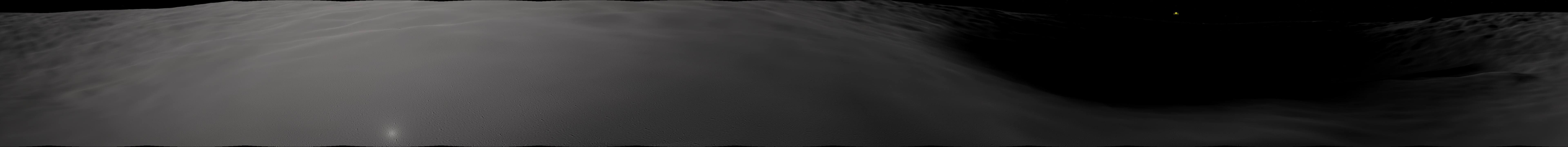}
        \caption{Panorama at the same location rendered with sun azimuth 90$^\circ$.}
    \end{subfigure}

    \vspace{6pt}

    \begin{subfigure}{\linewidth}
        \centering
        \includegraphics[width=\linewidth]{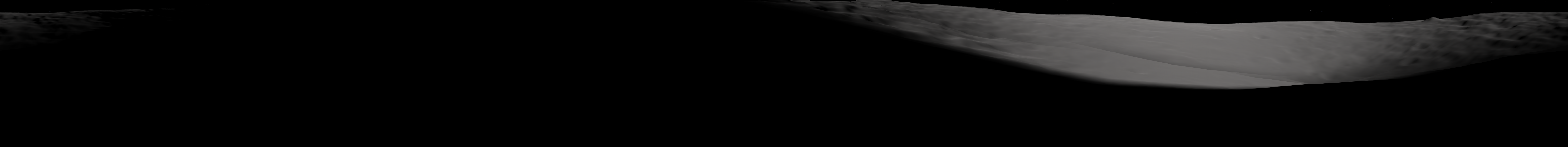}
        \caption{Panorama at the same location rendered with sun azimuth 270$^\circ$}
    \end{subfigure}

    \caption{Effect of changing sun azimuth on panorama appearance at the same location.}
    \label{fig:orientation}
\end{figure}

We conducted an additional experiment to investigate the robustness of this result with regards to inconsistent views.
For this purpose, we changed the sun azimuth from 90$^\circ$ to 270$^\circ$ in a separate render of the test set, while keeping sun elevation at 15$^\circ$, terrain geometry and POIs as before (see Fig.~\ref{fig:orientation} for an illustration). 
When evaluated on this new test set without re-training, TransGeo's performance drops dramatically to 4.04\% R@1 (14.72\% R@10), suggesting a high sensitivity to varying illumination conditions.

\subsection{Tile-based (VIGOR-style) retrieval}
\label{sec:results_vigor}

\begin{table}[t]
\caption{Retrieval accuracy for tile-based (VIGOR-style) data.}
\label{tab:results_vigor_lunar}

\begin{tabular}{lcccc}
\hline
Method & R@1 & R@5 & R@10 & R@1\% \\
\hline
TG mining        & 3.425 & 5.421 & 7.156 & 14.265 \\
Balancing only   & 7.365 & 17.556 & 28.904 & 38.409 \\
Ring mining      & 10.467 & 18.628 & 29.946 & 41.576 \\
VIGOR model      & 1.560 & 2.014 & 4.157 & 8.097 \\
\hline
\end{tabular}

\vspace{2mm}

\begin{tabular}{lcc}
\hline
Method & Hit@1 & Hit@5 \\
\hline
TG mining        & 8.726 & 15.260 \\
Balancing only   & 10.863 & 36.954 \\
Ring mining      & 13.924 & 39.066 \\
VIGOR model      & 3.011 & 5.072 \\
\hline
\end{tabular}
\end{table}

Table~\ref{tab:results_vigor_lunar} reports performance of TransGeo when applied to the tile-based variant of our dataset. 
We observe that the data mining strategy has a large impact on retrieval accuracy. 
In particular, the original TransGeo mining strategy leads to a substantial performance drop (R@1 = 3.43\%) when compared to data balancing (R@1 = 7.37\%). 

In contrast, our proposed ring-based mining introduces a simple geographic prior by sampling negatives from a ring-shaped region around the panorama location.
These negatives remain geographically close to the query but lie outside its positive and semi-positive tiles, yielding harder yet less ambiguous training examples in a more controlled way.

While the absolute performance of TransGeo is comparatively low, it still performs better than a na\"ive application of pre-trained baseline methods, such as the machine learning model proposed in the original VIGOR work~\cite{zhu2021vigor} which achieves only R@1 = 1.56\% (R@10 = 4.16\%). This suggests that models trained on urban datasets struggle to generalize well when confronted with planetary terrain.

\section{Discussion and Conclusion}

Our results indicate that modern machine learning models like TransGeo are, in principle, capable of learning meaningful cross-view correspondences for simulated planetary terrain. 
The performance gap between R@1 and Hit@1 suggests that, in many cases, the correct neighborhood of a panorama is retrieved, while the exact tile is not always ranked first. 
Given the inherent ambiguity of crater-dominated surfaces, this observation highlights the value of a tile-based dataset variant with semi-positive matches for characterizing localization performance. 
At the same time, it suggests that future work should investigate methods for improved fine-grained discrimination between neighboring tiles.

While we studied inconsistent illumination conditions only for the One-to-One variant, it is expected that tile-based retrieval would be similarly affected. 
Changing the sun azimuth by $\pm 180^\circ$ produced substantial appearance changes and led to a much larger degradation in accuracy than initially anticipated. 
It remains currently unclear whether similar breakdowns occur for smaller illumination changes (e.g., $\pm 5^\circ$) and whether data augmentation techniques, such as including different sun azimuth and sun elevations in the training data, could improve robustness.
Given that our ring mining strategy already improved retrieval accuracy in our experiments, a more systematic investigation of data mining and augmentation techniques appears to be a promising direction for future studies.

We acknowledge that our renderings have several limitations when compared to real imagery, such as that captured by the panoramic camera of the lunar Yutu rover\footnote{\url{https://planetary.s3.amazonaws.com/data/change3/pcam.html}}. 
In particular, our terrain textures can appear overly smooth and repetitive due to the limited availability of reliable small-scale surface detail. 
To better capture the visual complexity of real planetary imagery, we plan to further improve the diversity and richness of our dataset in the upcoming public release.

In conclusion, our work demonstrates that established cross-view localization techniques can be transferred to planetary terrain.
The dataset introduced in this work therefore provides a foundation for future research on vision-based localization for planetary exploration, with the long-term goal of enabling autonomous systems to achieve robust geo-localization and positional awareness during space missions.

\section*{Data Availability Statement}
This study was conducted on an evolving cross-view dataset and should therefore be regarded as a preliminary investigation. A public release of the dataset under the CC BY 4.0 license on Zenodo\footnote{\url{https://zenodo.org}} is planned for 2026, alongside a related scientific competition that will facilitate the benchmarking of novel cross-view localization techniques via an online leaderboard.

\printbibliography
\addcontentsline{toc}{section}{References}
\end{document}